\documentclass[journal,twoside,web]{ieeecolor}
\usepackage{generic}
\usepackage{cite}
\usepackage{bm}
\usepackage{amsmath,amssymb,amsfonts}
\usepackage{algorithmic}
\usepackage{graphicx}
\usepackage{algorithm,algorithmic}
\usepackage[hidelinks]{hyperref}
\usepackage{textcomp}
\usepackage{booktabs} 
\usepackage{multirow} 
\usepackage{lmodern}

\usepackage{makecell}

\def\BibTeX{{\rm B\kern-.05em{\sc i\kern-.025em b}\kern-.08em
 T\kern-.1667em\lower.7ex\hbox{E}\kern-.125emX}}
\begin{document}
\title{DrivenMorph: Bridging Attention Mechanism and Variational Image Registration via Difference Modeling}
\author{Mingke Li, Jianping Zhang, and Jinqiu Deng
\thanks{This work was partly supported by the science and technology innovation Program of Hunan Province (grant number 2024RC9008); and in part by the Project of Scientific Research Fund of the Hunan Provincial Science and Technology Department (grant number 2023GK2029); and in part by the Program for Science and Technology Innovative Research Team in Higher Educational Institutions of Hunan Province of China; and in part by the Postgraduate Scientific Research Innovation Project of Hunan Province under Grant (grant number CX20230632). Corresponding author: Jianping Zhang (jpzhang@xtu.edu.cn).}
\thanks{M. Li, J. Zhang, and J. Deng are with the School of Mathematics and Computational Science, Hunan Key Laboratory for Computation and Simulation in Science and Engineering, Xiangtan University, Xiangtan, Hunan 411105, P. R. China (e-mail: \{limingke$\_$xtu, jqdeng\}@smail.xtu.edu.cn, jpzhang@xtu.edu.cn).}}

\maketitle

\begin{abstract}
Medical image registration benefits significantly from deep learning, yet existing approaches often lack physical explainability and fine-grained deformation control. Motivated by Demons algorithms, we propose a novel DrivenMorph framework that bridges attention mechanisms with variational image registration by incorporating difference modeling as a physically inspired inductive bias. The resulting driving force, computed from local differences in the latent feature space, provides explicit semantic guidance throughout the registration process. It directly drives the registration process through a neural Demons layer that simulates force-displacement interactions to generate smooth and anatomically consistent deformation. Unlike previous methods, our approach not only integrates traditional registration principles with popular deep networks, providing an explainable and efficient solution for learning-based medical image registration, but also separates difference modeling from deformation, improving modularity and explainability. Extensive experiments on multiple 3D brain MRI datasets demonstrate superior performance over state-of-the-art learning-based and optimization-based methods. Furthermore, visualizations and statistical analyses confirm that the learned driving force aligns closely with actual deformation patterns, supporting its explanatory value. 
\end{abstract}


\section{Introduction}
\label{sec:introduction}
\IEEEPARstart{D}{eformable} image registration is crucial to align anatomical structures under different conditions or between individuals. By establishing smooth and non-linear correspondences between fixed and moving images, it allows accurate anatomical alignment to facilitate tasks such as tumor tracking and surgical navigation \cite{tu2024coarse, baptista2024keypoint, wu2024noise, dong2023preserving}. It also serves as a foundational step for downstream workflows such as segmentation, image fusion, and radiomic analysis \cite{castadotAssessmentDeformableRegistration2010, ding2024image, yadav2025modified, li2025cyclic}. However, in practical clinical environments, registration algorithms are required to be not only accurate and robust, but also structurally transparent, allowing users to inspect the decision-making process.

Traditional registration techniques modeled by manually designed features or energy minimization are both interpretable and theoretically founded \cite{wen2019incorporation}, but face difficulties with complex deformations and can be computationally expensive \cite{chen2025survey}. Recent developments in convolutional-based \cite{he2016deep,dalcaUnsupervisedLearningProbabilistic2019a,balakrishnanUnsupervisedLearningModel2018a,feng2024weakly} and transformer-based \cite{dosovitskiy2021an, liu2021swin, Transmorph,chenViTVNetVisionTransformer2021,shi2022xmorpher} architectures have provided a new perspective to solve the complex registration problem. These models extract high-dimensional features to improve anatomical correspondence and reduce inference time compared to iterative optimization-based methods.

\begin{figure}[htbp]
\centering
\includegraphics[width=\linewidth]{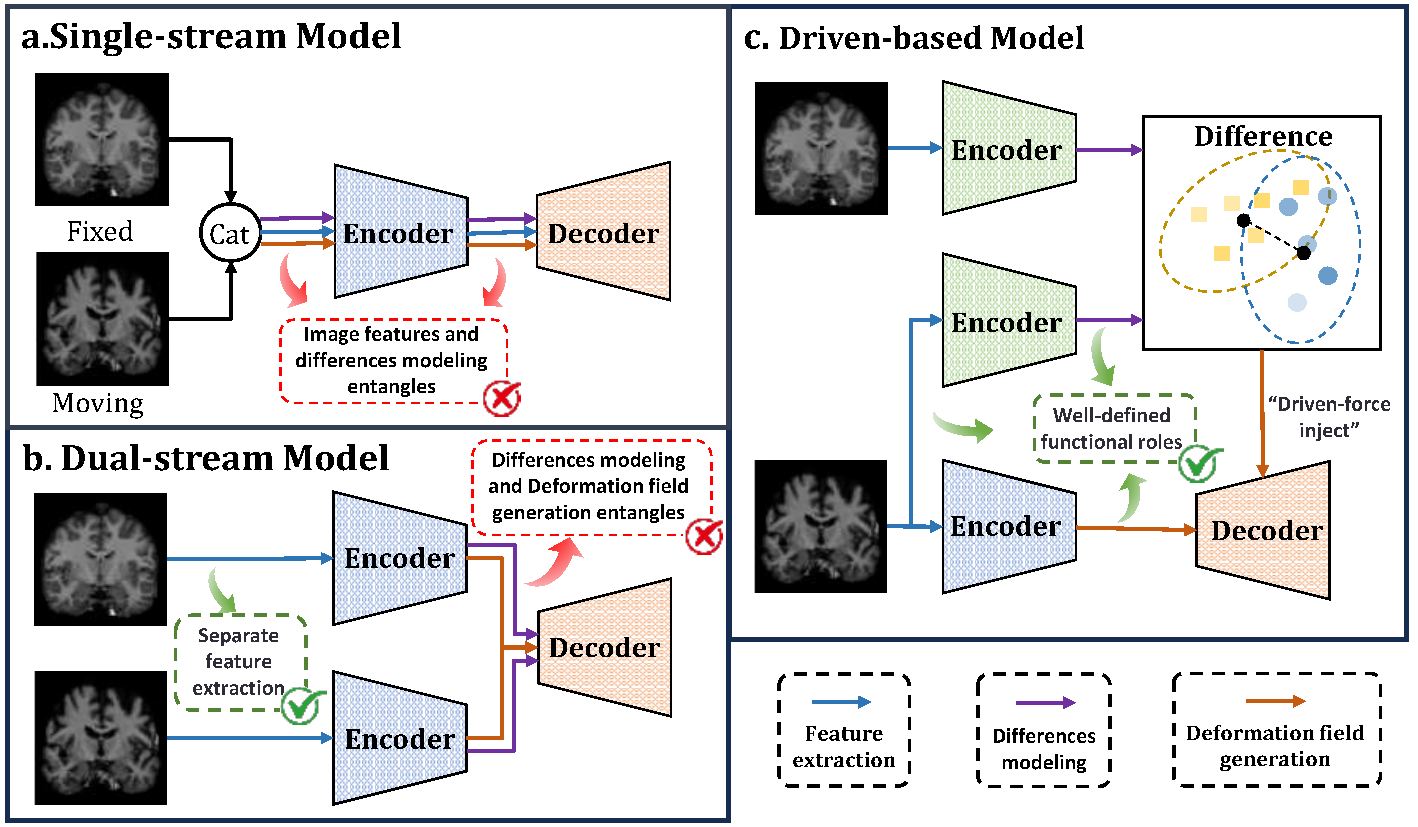}
\caption{Comparison of the proposed driven-based architecture with single-stream and dual-stream architectures.}
\label{fig1_1}
\end{figure}

Early deep learning registration methods often adopt a single-stream architecture (Fig. \ref{fig1_1} (a)) \cite{balakrishnanVoxelmorphLearningFramework2019, dalca2019learning, li2025cyclic}, where fixed and moving images ($I_f$, $I_m$) are concatenated and jointly processed by a single encoder. This architecture requires the encoder to simultaneously extract semantic features and capture inter-image differences, but lacks explicit comparison mechanisms found in Siamese networks \cite{koch2015siamese, bertinetto2016fully, radford2021learning}. Consequently, intra-image representation and inter-image discrepancy modeling become entangled, complicating learning, and weakening interpretability.

To overcome these limitations, recent work has implemented a dual-stream architecture (Fig. \ref{fig1_1}(b)) \cite{chen2023transmatch, yang2025tcde, kang2022dual, wu2024saliency}. This design uses independent encoders to extract the features from $I_f$ and $I_m$ separately, reducing the entanglement of the features and focusing each stream on learning the representation of a single image. The decoder then fuses these features to explicitly model inter-image correspondences. This architecture improves training flexibility, facilitates the integration of more robust pre-trained backbones \cite{gui2024survey, chen2020simple}, and improves overall performance.

However, current dual-stream architectures typically bypass explicit difference modeling by directly decoding the fused features into a deformation field, overlooking a critical insight from classical variational registration: image differences should act as driving forces rather than be directly mapped to deformation. Furthermore, although the encoder is decoupled, the decoder still entangles difference modeling and deformation generation, blurring functional roles and weakening interpretability. Inspired by Demons-based methods \cite{thirion1998image}, we argue that explicitly modeling image differences as intermediate driving forces provides a more structured grounded pathway to deformation, potentially improving model stability and generalization.

To address structural and functional entanglement in dual-stream decoders, we propose a decoupled design that separates difference modeling from deformation mechanism. This enables a driving force registration strategy, where feature-space dissimilarities are transformed into explicit deformation cues, as shown in Fig. \ref{fig1_1}(c). We also design the Neural Demons Layer, which simulates the classical “force × direction” paradigm to guide the decoding. Embedded in a multi-scale pyramid, our framework delivers more transparent workflows, improved structural interpretability, and flexible deformation modeling. The main contributions of this work are summarized as follows:
\begin{itemize}
\item We propose a novel DrivenMorph framework for medical image registration. Inspired by the structural decoupling in the Demons algorithm, it can be explicitly separated into feature extraction, difference modeling, and deformation decoding, thus introducing inductive bias and improving structural interpretability.

\item We develop a difference modeling method based on local latent feature dissimilarity to provide explicit semantic guidance for the learnable registration.

\item We design a Neural Demons Layer to link the attention mechanism with variational image registration. It employs cross-attention and self-attention to transform driving force fields into deformation forces: cross-attention injects semantic discrepancies into the deformation space, whereas self-attention enforces spatial coherence, together simulating the direction and magnitude-based deformation mechanism of classical variational Demons.
    
\item We introduce a deep supervised loss to constrain difference modeling, ensuring the accuracy of driving force estimation and improving registration stability.
\end{itemize}

\section{Related Work}\label{sect02}
\subsection{The Demons Algorithm in Variational Image Registration}
The variational image registration algorithms aim to find a spatial transformation ${\bm{\phi}}: \Omega \to \mathbb{R} ^d $ that aligns a moving image $I_m$ with a fixed image $I_f$ by minimizing an optimization problem:
\begin{equation}
	{\bm{\phi}}^* = \arg \min_{{\bm{\phi}}} \mathcal{S}(I_f, I_m \circ {\bm{\phi}}) + \lambda \mathcal{R}({\bm{\phi}}),
	\label{eq2.1}
\end{equation}
where $\mathcal{S}(\cdot , \cdot)$ denotes a similarity metric (e.g. SSD, NCC), while $\mathcal{R}(\cdot)$ serves as a regularizer \cite{lorenzi2013lcc} enforcing smoothness and invertibility. The scalar $\lambda$ balances the trade-off between data fidelity and deformation regularity.

Among non-parametric methods, the Demons algorithm introduced by Thirion \cite{thirion1998image} is particularly popular because of its simplicity and computational efficiency. In its standard formulation, Demons refines the deformation field through iterative local updates driven by image information, which reduces the computational burden associated with full variational optimization.
At each iteration $\bm{\phi}=\bm{\phi}+\bm{u}$, the displacement field $\bm{u}$ is updated based on local image information and can be formulated as:
\begin{equation}
	\bm{u}(\bm{x}) = \underbrace{\frac{\delta I(\bm{x})}{\|\bm{g}(\bm{x})\|^2 + \eta(\bm{x})}}_{\text{magnitude}} \cdot \underbrace{\bm{g}(\bm{x})}_{\text{direction}},
	\label{e2.1}
\end{equation}
where $\delta I(\bm{x})$ represents the inter-image discrepancy (e.g. $I_f-I_m$), $\bm{g}(\bm{x})$ denotes a direction vector that guides the update (e.g. intra-image discrepancy $\bm{g}=\nabla I_m$), and $\eta(\bm{x})$ is a local regularization or uncertainty term that stabilizes the update. This formulation reflects a common physical interpretation: the displacement is modulated by the mismatch between images and driven by local intra-image structural information.

Demons-based registrations have explored increasingly sophisticated formulations of the driving force. Vercauteren et al. \cite{vercauteren2008symmetric} introduced a symmetric \emph{log}-Demons to enforce inverse consistency by applying equalized driving forces to both $I_m$ and $I_f$. To improve the precision of deformation, Zhang and Wen \cite{zhang2016log} proposed a modified \emph{log}-Demons framework with an explicitly constructed directional force, while Wen et al. \cite{wen2019incorporation} further incorporated structural tensor fields to guide the force along local anatomical orientations, achieving improved alignment near structural boundaries and within complex tissues.

These works reflect a clear progression from symmetric geometric modeling to structure-aware force adaptation, leading to improved registration accuracy and robustness. However, most Demons-based formulations still rely on hand-crafted driving forces derived from image gradients or structural priors. Although effective in specific scenarios, such heuristic designs struggle to capture the context-dependent nature of complex anatomical deformations. To address this, we explore the integration of the Demons algorithm into deep learning as an inductive bias, enabling models to learn  meaningful deformation patterns and enhancing the inspectability of the transformation mappings.

\subsection{Deep Learning Method}
Learning-based registration methods have gained increasing attention for their superior speed and accuracy. Unlike traditional methods that rely on iterative optimization of energy functions, deep learning models adopt an end-to-end approach that directly maps image pairs to deformation fields, eliminating the need for iterative inference and enabling real-time registration \cite{XJia2025a}.

VoxelMorph \cite{balakrishnanVoxelmorphLearningFramework2019} introduced the first unsupervised deep learning framework for image registration, where similarity metrics and regularization terms are optimized jointly within a neural network. By mirroring the formulation of traditional energy-based models, it enabled end-to-end training without requiring ground-truth deformation fields. Building on this foundation, subsequent studies explored architectural innovations, loss function refinements, and enhanced feature representations to further improve performance.
Among these, TransMorph \cite{Transmorph} replaced CNNs with transformer architectures, demonstrating their effectiveness through comprehensive evaluations. Similarly, recent studies such as XMorpher\cite{shi2022xmorpher} and TransMatch\cite{chen2023transmatch} have employed cross-attention mechanisms to improve the integration of features from moving and fixed images. Nonetheless, despite these architectural refinements, these approaches essentially remain within the framework of directly regressing deformation fields from intertwined feature representations. As a result, they frequently lack the explicit physical driving mechanism and domain-specific inductive biases that are intrinsic to variational registration theories.

In contrast, several recent studies have embedded domain-specific constraints to improve deformation modeling \cite{WJHuang2021,tian2024nephi,KChen2025a}. Huang et al. \cite{WJHuang2021} designed an end-to-end coarse-to-fine network architecture consisting of a dual consistency constraint and a prior knowledge-based loss function to enhance the registration performances. Nephi \cite{tian2024nephi} advanced diffeomorphic modeling using continuous implicit representations and neural ODEs to obtain smooth and invertible transformations. Chen et al. \cite{KChen2025a} randomly generated Beltrami coefficients to produce a series of diffeomorphic labels for training the supervised few-shot learning network. Our method follows this philosophy by integrating prior assumptions from the classical Demons algorithm into network design, achieving principled deformation modeling while retaining the advantages of deep learning.

\subsection{Difference Modeling}
We aim to induce the deformation field through difference modeling, which is inspired by metric learning \cite{bromley1993signature, koch2015siamese, chen2017beyond}. Difference modeling inherently evaluates similarity, aligning closely with the aims of medical image registration. However, this perspective remains underexplored, as many registration models rely on simple feature concatenation without explicitly modeling task-specific differences.

Representative frameworks such as Siamese networks \cite{bromley1993signature} compare paired inputs by minimizing intra-class distances and maximizing inter-class ones, laying the foundation for deep metric learning. Variants such as one-shot Siamese models \cite{koch2015siamese} and quadruplet-loss formulations \cite{chen2017beyond} further enhance discriminative power, particularly in data-scarce settings. Contrastive learning extends this paradigm to unsupervised representation learning, making it well-suited for medical image registration where annotations are limited.

Despite its potential, existing contrastive learning methods, such as CLIP \cite{radford2021learning} and SimCLR \cite{chen2020simple}, mainly target global representations, which are difficult to align with the fine-grained local spatial consistency required in registration. Bridging contrastive learning with local correspondence modeling thus remains an open challenge.

This work introduces contrastive learning into the difference modeling, extending traditional Demons formulations beyond pixel-level gradients into deep feature space. Using local feature dissimilarity, the model captures richer structural variations and provides a more expressive and interpretable deformation representation.

\section{Methodology}\label{sect03}
Let the fixed volume $I_f$ and the moving volume $I_m$ be two input volumes defined in the 3D space $\Omega \subseteq \mathbb{R}^3$.The registration problem involves determining a deformation field ${\bm{\phi}} \in \mathbb{R}^{H \times W \times D \times 3}$ to align $I_m$ with $I_f$. Our main goal is to achieve the spatial alignment of the corresponding anatomical structures between $I_m$ and $I_f$.

\begin{figure*}[htbp]
\centering
\includegraphics[width=\linewidth]{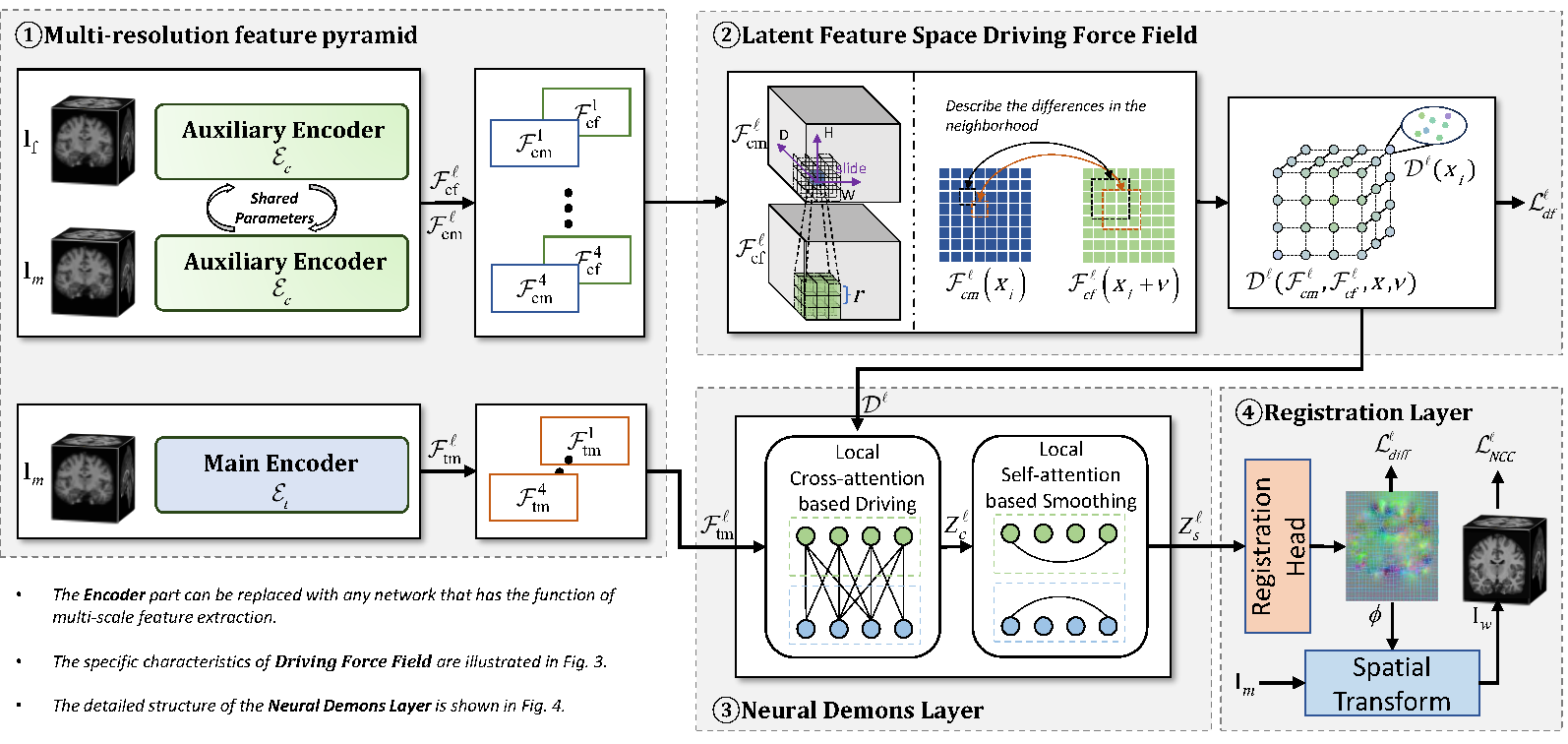}
\caption{Overall architecture of DrivenMorph. The driving force field is constructed by computing the paired features 
$[\mathcal{F}^\ell_{cf},\mathcal{F}^\ell_{cm}]$, which encode differential information for the Neural Demons Layer. This layer leverages the driving force to guide the transformation of the moving image features $\mathcal{F}^\ell_{tm}$ within the feature space. The resulting features are then decoded in the Registration Layer to generate a deformation field, from which the warped image $I_w$ is obtained via spatial sampling.}
	\label{fig3_1}
\end{figure*}

\subsection{Multi-resolution Feature Pyramid Encoder}
Our encoder architecture consists of a main encoder based on transformers and two auxiliary encoders based on CNNs. 
Specifically, the main encoder employs the Swin Transformer \cite{liu2021swin} (embedding dimension $C=24$, heads $\{4,4,8,8\}$, window sizes $\{5,6,7\}$). Its self-attention mechanism supports effective modeling of long-range dependencies across the entire image, thereby providing a robust basis for deformation learning. The auxiliary encoder is constructed using ResNet-RS\cite{bello2021revisiting} (channels $\{24, 48, 96, 192\}$, blocks $\{3,3,8,4\}$) and is designed to encode latent feature representations of the driving force field. In contrast to the main encoder, the auxiliary encoder emphasizes local information. Motivated by Demons algorithms, where driving forces are determined by local intensity variations within a restricted neighborhood, we adopt a convolution-based design. This CNN architecture naturally encodes a locality-inducing inductive bias, which makes it particularly responsive to local structures\cite{elguendouze2023explainability} and consistent with our neighborhood correlation formulation given in Eq.\eqref{driving force}.

To improve computational efficiency, an early downsampling block is applied at the input of both encoders. With this setup, most of the computationally demanding operations occur at lower spatial resolutions, and the resulting deformation field is then upsampled back to the original full resolution. Following this preprocessing, each encoder independently extracts a hierarchy of multi-scale features from the input images. The main encoder $\mathcal{E}_t$ extracts only features from $I_m$, while the auxiliary encoder $\mathcal{E}_c$ processes both the moving image $I_m$ and the fixed image $I_f$ using a shared weight CNN in a dual-stream manner \cite{kang2022dual}. 
As shown in Fig.\ref{fig3_1}, the resulting latent feature maps are defined as
\begin{equation}
	%\footnotesize
	\left\{
	\begin{aligned}
		\mathcal{F}^\ell_{tm} &= \mathcal{E}^\ell_\text{t}(I_m) \\
		\mathcal{F}^\ell_{cm} &= \mathcal{E}^\ell_\text{c}(I_m)\\
		\mathcal{F}^\ell_{cf} &= \mathcal{E}^\ell_\text{c}(I_f)
	\end{aligned}
	\right\}
	\in \mathbb{R}^{\frac{H}{2^\ell} \times \frac{W}{2^\ell} \times \frac{D}{2^\ell} \times C^{\ell}}, \quad \ell = 1, 2, 3, 4
	\label{eq3.1}
\end{equation}
where $\mathcal{F}^\ell_{tm}$ denotes the features of $I_m$ extracted by the Transformer-based encoder, while $\mathcal{F}^\ell_{cm}$ and $\mathcal{F}^\ell_{cf}$ represent the features of $I_m$ and $I_f$ extracted by the CNN-based auxiliary encoder. These hierarchical features form the basis for the subsequent modules, including cross-scale fusion and neighborhood descriptor computation.

The use of shared weights in the auxiliary encoder is motivated by three considerations:
\begin{itemize}
	\item $I_m$ and $I_f$ originate from the same modality or possess very similar anatomical structures as well as local textures;
	\item Embedding both images into a unified latent feature space \cite{radford2021learning} facilitates similarity-based matching and the learning of consistent deformation fields;
	\item This design echoes the symmetry of the classical Demons algorithm \cite{vercauteren2008symmetric}, improving the interpretability and training stability.
\end{itemize}

\subsection{Latent Feature Space Driving Force Field}
In registration tasks, it is essential to analyze both images to identify and capture their structural differences. The goal of registration is to establish transformations that maximize alignment similarity by these differences. The Demons registration framework \cite{wen2019incorporation} evaluates local intensity variations or gradient information at each pixel to construct a driving force field. Inspired by this algorithm, we propose a driving force field module that incorporates a latent feature space at each coordinate to represent the feature differences between $I_m$ and $I_f$. Such latent feature differences can be used to compute contextual feature descriptors for quantifying inter-image differences.

Contrastive learning has been widely adopted to map data into a unified embedding space, aiding matching and retrieval tasks via similarity-driven optimization. In our scenario, this matching mechanism is similar to the driving force in image registration, both of which involve evaluating distances or similarities between feature representations. Thus, we begin with correlation matrices to determine the driving force.

We compute the feature differences between $\mathcal{F}^\ell_{cm}, \mathcal{F}^\ell_{cf} \in \mathbb{R}^{h \times w \times d \times c}$, where $h$, $w$, $d$, and $c$ denote the spatial dimensions and the number of channels in the feature map at layer $\ell$. Since image registration aims to achieve spatial alignment between image pairs, it is also expected that their corresponding feature maps are spatially aligned \cite{teed2020raft}. To evaluate the feature similarity at each pixel in $\mathcal{F}^\ell_{cm}$, we compute the full correlation matrix \cite{wang2020learning} of $\mathcal{F}^\ell_{cm}$ with all pixels in $\mathcal{F}^\ell_{cf}$ as follows:
\begin{equation}
	Corr_{fc}^\ell(\mathcal{F}^\ell_{cm}, \mathcal{F}^\ell_{cf})=\frac{(\mathcal{F}^\ell_{cm})^{T} F^\ell_{cf}}{\sqrt{c}}\in\mathbb{R}^{(h w d)\times (h w d)},
	\label{global corr}
\end{equation}

where each element in the correlation matrix $Corr_{fc}^\ell$ corresponds to the similarity between a spatial position $\bm{x}\in\mathbb{R}^3$ in $\mathcal{F}^\ell_{cm}$ and $\bm{x}'\in\mathbb{R}^3$ in $\mathcal{F}^\ell_{cf}$. To prevent excessively large values from the dot product operation, we apply a scaling factor of $\sqrt{c}$ \cite{vaswani2017attention}. 

\begin{figure}[htbp]
	\centerline{\includegraphics[width=\linewidth]{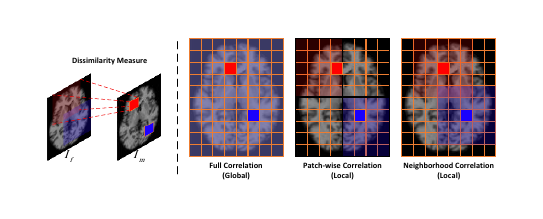}}
	\caption{Illustrations of different correlation characteristics.}
	\label{fig3_3}
\end{figure}

As illustrated in Table~\ref{tab:Flops}, full correlation substantially increases memory and computational requirements quadratically with spatial resolution. This method requires $2c \cdot (hwd)^2$ floating-point operations and tracks $(hwd)^2$ similarity values, resulting in impractical costs even for moderate feature map sizes. Importantly, as indicated in Fig.~\ref{fig3_3}, full correlation ignores the spatial locality assumption crucial in most image registration tasks, which depend on local geometric consistency and neighborhood-level feature alignment to determine deformation fields. These observations motivate the development of a localized correlation strategy aimed at reducing computational demands while preserving the adequate discriminative capacity.

\begin{table}[h]
	\centering
	%\scriptsize
    \caption{Computational cost and memory usage of different correlation strategies.}
	\label{tab:Flops}
	\begin{tabular}{lcc}
		\toprule
		\textbf{Module} & \textbf{FLOPs} & \textbf{Memory} \\
        & \textbf{(approx.)} & \textbf{(elements)} \\
		\midrule
		Full($Corr_{fc}$) & $2c \cdot (hwd)^2$ & $(hwd)^2$ \\
		Patch-based ($Corr_p$) & $2c \cdot p^3 \cdot hwd$ & $p^3 \cdot hwd$ \\
		Neighborhood ($Corr_{nh}$) & $2c \cdot (2r+1)^3 \cdot hwd$ & $ (2r+1)^3 \cdot hwd$ \\
		\bottomrule
	\end{tabular}
\end{table}

To reduce the computational burden of full correlation, similarity is assessed within local patches by dividing the feature volume into fixed-size blocks and performing correlation computations limited to each patch. This approach significantly reduces complexity, as indicated in Table~\ref{tab:Flops}.

However, the fundamental limitation of this approach is its inability to encode explicit directional correspondence. The patch-based correlation ($Corr_{p}$) can be interpreted as a similarity operator that is defined over a fixed absolute domain:
\begin{equation}
	Corr_{p}(\bm{x}, \bm{y}) = (\mathcal{F}_{cm}(\bm{x}))^T \mathcal{F}_{cf}(\bm{y}), \quad \forall \bm{x}, \bm{y} \in \mathcal{P}_k ,
\end{equation}
where $\mathcal{P}_k$ denotes the $k$-th non-overlapping spatial patch. In this formulation, every location $\bm{x}$ within the same patch $\mathcal{P}_k$ is correlated with an identical set of locations $\bm{y}$, regardless of its relative position inside the patch.
	
As a result, the computed correlations encode only the magnitude of local similarity but do not preserve an explicit association between similarity responses and relative spatial displacements $\bm{y}-\bm{x}$. This absence of relative spatial displacements prevents $Corr_{p}$ from providing directional information, which is essential for force-based deformation updates.

This limitation motivates the adoption of a sliding local neighborhood correlation, explicitly designed to encode displacement-aware correspondences at every spatial position. To construct a driving force field at layer $\ell$ that captures both feature similarity and relative spatial geometry, we introduce the neighborhood correlation in the latent feature space as:
\begin{equation}
	Corr_{nh}^\ell(\mathcal{F}^\ell_{cm}, \mathcal{F}^\ell_{cf}, \bm{x}, \bm{y})=(\mathcal{F}^\ell_{cm}(\bm{x}))^T \mathcal{F}^\ell_{cf}(\bm{y})\in \mathbb{R},
	\label{local_corr}
\end{equation}
where $\bm{x} \in \mathbb{R}^3$ and $\bm{y} \in \Omega_r(\bm{x}) := \{ \bm{x} + \bm{\nu} \mid \|\bm{\nu}\|_\infty \le r \}$ denote spatial locations in the feature volumes $\mathcal{F}^\ell_{cm}$ and $\mathcal{F}^\ell_{cf}$, respectively. Here, $\bm{\nu} \in \mathbb{Z}^3$ represents the displacement vector relative to $\bm{x}$, restricting the correlation computation to a local neighborhood with a fixed-radius. 

In contrast to the patch-based correlation $Corr_p$, which is computed over absolute patch domains, our formulation defines the neighborhood $\Omega_r(\bm{x})$ relative to the central voxel $\bm{x}$, thereby introducing a local coordinate frame at every spatial position. This construction ensures that each similarity response is explicitly associated with a distinct relative displacement $\bm{\nu} = \bm{y}-\bm{x}$. As a result, the correlation tensor $\mathcal{D}^\ell(\bm{x}) \in \mathbb{R}^{(2r+1)^3}$ is no longer just a set of unordered similarity values, but rather a displacement-indexed descriptor whose channel arrangement reflects the spatial directions within the local neighborhood.
	
This displacement-aware structure enables the Neural Demons Layer to interpret correlation responses as directionally consistent driving signals, which is essential for force-based deformation updates in Demons-style registration frameworks.

Using the neighborhood correlation, the driving force field $\mathcal{D}^\ell$ between $\mathcal{F}^\ell_{cm}$ and $\mathcal{F}^\ell_{cf}$ is defined as follows:

\begin{equation}
	\mathcal{D}^\ell(\mathcal{F}^\ell_{cm}, \mathcal{F}^\ell_{cf}, \bm{x}, \bm{y})=-\frac{Corr_{nh}^\ell(\mathcal{F}^\ell_{cm}, \mathcal{F}^\ell_{cf}, \bm{x}, \bm{y})}{\left\|\mathcal{F}^\ell_{cm}(\bm{x}) \right\|_2 \left\|\mathcal{F}^\ell_{cf}(\bm{y}) \right\|_2 },
	\label{driving force}
\end{equation}
where normalization yields a measure of cosine similarity within the local neighborhood $\Omega_r(\bm{x})$. Crucially, the resulting tensor $\mathcal{D}^\ell \in \mathbb{R}^{h \times w \times d \times (2r+1)^3}$ functions as a local displacement-indexed driving volume for all voxels in $\mathcal{F}^\ell_{cm}$. Each channel along the last dimension corresponds strictly to a specific relative displacement $\bm{\nu}$.

Because neighboring spatial positions have strongly overlapping correlation neighborhoods, the driving force descriptor $\mathcal{D}^\ell(\bm{x}) \in \mathbb{R}^{(2r+1)^3}$ changes smoothly as a function of $\bm{x}$ while maintaining a consistent ordering of displacements. This design yields a force signal that is both spatially coherent and directionally aligned, which is crucial for the Neural Demons Layer to carry out force-guided deformation updates in a Demons-like fashion. As shown in Table \ref{tab:Flops}, this formulation significantly reduces computational and memory complexity compared to full correlation, while retaining the locality and spatial consistency required for dense deformable registration.

Our design is inspired by the formulation of driving force in the classical Demons algorithm, where movement is driven by local intensity differences and gradient-based forces. In contrast, our approach forms driving forces within the latent feature space, utilizing projected neighborhood correlations instead of direct intensity differences. These descriptors, based on contextual feature dissimilarity, maintain geometric and semantic interpretability, improving their effectiveness in guiding deformation estimation within learning-based registration frameworks.

\subsection{Neural Demons Layer}
\subsubsection{Local Cross-attention based Driving (LCBD)}
In the previous section, we introduced the driving force field $\mathcal{D}^\ell$ and the feature map $\mathcal{F}^\ell_{tm}$. Building on these, we propose a locally constrained magnitude mechanism for displacement fields. Here, each coordinate $\bm{x} \in \Omega$ in $\mathcal{F}^\ell_{tm}$ is driven solely by its corresponding $\mathcal{D}^\ell(\bm{x})$ and its spatial neighborhood.
This design follows the principle of the Demons algorithm, where the deformation \eqref{e2.1} results from the product of a local magnitude and direction component. Inspired by this strategy, we develop a learnable local update in the feature space. Thus, the computation at a point $\bm{x}$ is defined as
\begin{equation}
	Z_c(\bm{x}) = \underbrace{\text{softmax}\left( \frac{A_q(\bm{x})}{\sqrt{m}} \right)}_{\text{magnitude}} \underbrace{V_q(\bm{x})}_{\text{direction}},
	\label{e3.3}
\end{equation}
where softmax weights act as an adaptive magnitude modulator and $V_q(\bm{x})$ provides direction guidance from the local structure. This formulation maintains the separation of scalar and vector terms while permitting spatially variable and data-driven updates. Here, $q=(2r+1)^3$ represents the total points in a local neighborhood $\Omega_{r}(\bm{x})$ centered at $\bm{x}$. We enumerate these neighbors as $\{\rho_1(\bm{x}), \rho_2(\bm{x}), \ldots, \rho_q(\bm{x}) | \rho_j(\bm{x}) \in \Omega_{r},j=1, \ldots, q\}$. 

It is evident from \eqref{e3.3} that the local deformation of $Z_c(\bm{x})$ evolves into a learnable variant of the Demons algorithm (\ref{e2.1}), which incorporates both difference modeling and a structured spatial prior via $\Omega_{r}(\bm{x})$. Next, we will provide descriptions for $\text{softmax}\left( \frac{A_q(\bm{x})}{\sqrt{m}} \right)$ and $V_q(\bm{x})$.

\begin{figure*}[htbp]
	\centerline{\includegraphics[width=0.75\linewidth]{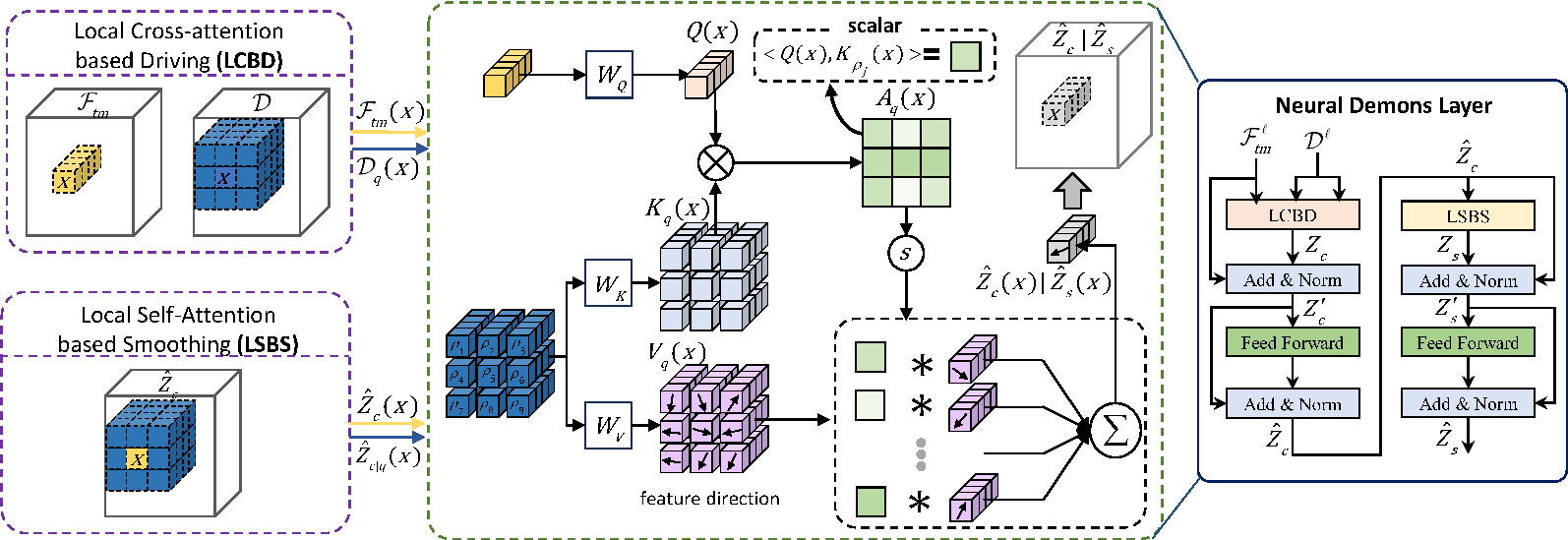}}
	\caption{The overview of the Neural Demons Layer includes Local Cross-attention based Driving (LCBD) and Local Self-Attention based Smoothing (LSBS).}
	\label{fig3_4}
\end{figure*}

\paragraph{$\text{softmax}\left( \frac{A_q(x)}{\sqrt{m}} \right)$ term} 
In the classical Demons algorithm, the magnitude of the deformation update is modulated by the residual intensity difference $I_f(\bm{x})-I_m(\bm{x})$, which quantifies the necessary correction at the location $x$.
We reinterpret this residual as a feature-space dissimilarity, which is computed via the inner product between an embedded query $Q(\bm{x})$ and a set of key vectors $K_q(\bm{x})$ from its neighborhood $\Omega_{r}$, i.e.,
\begin{equation}
\begin{split}
sim_{\rho_j}(\bm{x})=Q(\bm{x}) \cdot &K_{\rho_j(\bm{x})}^T,\quad \forall\rho_j(\bm{x})\in \Omega_{r},\\
Q=\mathcal{F}_{tm}&W_Q\in\mathbb{R}^{h \times w \times d \times c},\\K=\mathcal{D}&W_K\in\mathbb{R}^{h \times w \times d \times c},
	\label{e3.5}
\end{split}
\end{equation}
where $W_Q$ and $W_K$ are two learnable linear projections, respectively.

Accordingly, we define the pre-softmax attention weight vector $A_q(\bm{x}) \in \mathbb{R}^{q}$ as:
\begin{equation}
	A_q(\bm{x}) = \begin{bmatrix}
		Q(\bm{x}) K_{\rho_1(\bm{x})}^T + b_{\rho_1(\bm{x})} \\
		Q(\bm{x}) K_{\rho_2(\bm{x})}^T + b_{\rho_2(\bm{x})} \\
		\vdots \\
		Q(\bm{x}) K_{\rho_q(\bm{x})}^T + b_{\rho_q(\bm{x})}
	\end{bmatrix},
	\label{e3.6}
\end{equation}
where $b_{\rho_j(\bm{x})}$ encodes the relative positional bias between the coordinate $x$ and its neighbor $\rho_j(\bm{x})\in \Omega_{r}(\bm{x})$. This term is crucial for maintaining the spatial directionality of the interactions, which would otherwise be lost due to the flattening operations employed in the attention computation.
We can also see that $A_q(\bm{x})$ represents a set of similarity scores between the central point $\bm{x}$ and its neighboring locations $\rho_j(\bm{x})\in\Omega_{r}$. After applying softmax normalization, these scores reflect each neighbor's relative influence on deformation updates. Similarly to the residual term in the classic Demons method, $A_q(\bm{x})$ serves as the magnitude component in the learned update.

\paragraph{\texorpdfstring{$V_q$}{Vq} term}
The directional component $V_q$ is obtained by applying a linear projection to $\mathcal{D}$, that is, $V=\mathcal{D}W_V$. The matrix of neighborhood values, $V_q(\bm{x})$, for the $\rho_j(\bm{x})$-th feature point, is constructed by arranging its $q$ nearest neighbor value vectors
\[V_q(\bm{x})=[{V_{\rho_1(\bm{x})}^T}, {V_{\rho_2(\bm{x})}^T}, \dots, {V_{\rho_q(\bm{x})}^T}]^T,\]
where $V_q(\bm{x})$ encodes driving force field $\mathcal{D}$ around coordinate $\bm{x}$ in a latent form that captures the directional features within the neighborhood $\Omega_{r}(\bm{x})$. This representation indicates potential deformation directions and acts as a directional basis in weighted aggregation. Similarly to the image gradient $\nabla I_f(\bm{x})$ in the original Demons algorithm, $V_q(\bm{x})$ serves as a learned direction field refined by local similarity patterns.

The above operations are independently applied at each spatial coordinate, yielding the final feature map: 
\begin{equation}
	Z_c = \{ Z_c(\bm{x})=\text{softmax}\left( \frac{A_q(\bm{x})}{\sqrt{m}} \right) V_q(\bm{x})|\bm{x} \in \Omega\}.
	\label{e3.7}
\end{equation}
After computing the attention output $Z_c$, we apply a residual connection with the query input $\mathcal{F}_{tm}$, followed by a layer normalization operation. The resulting features are then passed through a feedforward network (FFN) with another residual connection and layer normalization step\cite{sharma2025ilam}. The overall local cross-attention based driving (LCBD) process can be described as
\[\begin{split}
	Z_c & =LCBD(\mathcal{F}_{tm},\mathcal{D}),\\
	Z'_c & = LayerNorm(Z_c+\mathcal{F}_{tm}), \\
	\hat{Z_c}& =LayerNorm(FFN(Z'_c)+Z'_c).
\end{split}\]

\subsubsection{Local Self-attention based Smoothing (LSBS)}
In the Demons algorithm \cite{pennec1999understanding}, Gaussian smoothing is applied to the displacement updates to suppress local irregularities and stabilize the optimization. This insight motivates the design of a mechanism that encourages local consistency in learned deformation representations. In our framework, however, the deformation field is not directly optimized; instead, it is incrementally formed via learned feature transformations. Consequently, a basic smoothing operator cannot adequately transform the guided deformation features generated by LCBD into expressive deformation updates. We therefore require a learnable module that simultaneously maintains local spatial consistency and offers sufficient transformation capacity for the deformation features.

To this end, we introduce the Local Self-attention based Smoothing (LSBS) module, which operates on the fused deformation features $\hat{Z}_c$. The primary difference between LSBS and the previous LCBD module lies in the source of their attention components. In LCBD, the keys and values are taken from the latent feature driving force field, allowing the deformation directions to be guided by feature discrepancies. In contrast, LSBS generates all queries, keys, and values directly from the deformation feature itself:
\begin{equation}
	Q=\hat{Z}_cW_Q, K=\hat{Z}_cW_k, V=\hat{Z}_cW_V.
	\label{e3.8}
\end{equation}

This modification in structure redefines the function of the attention mechanism. Instead of introducing external guidance signals, LSBS applies a learnable local transformation to the deformation features that have already been guided by LCBD. From an algorithmic standpoint, LSBS can be interpreted as a learnable replacement of the Gaussian smoothing step typically used in classical registration approaches. In particular, the update in \eqref{e3.3} carries out a weighted aggregation within a local neighborhood:
\begin{equation}
	Z_s(\bm{x})=\sum_{j=1}^{q}\text{softmax}\left(\frac{A_j(\bm{x})}{\sqrt{m}}\right) \cdot V_{\rho _j(\bm{x})},
	\label{e3.9}
\end{equation}
which is functionally equivalent to a local smoothing operation.

In contrast to Gaussian filtering, which uses fixed, isotropic weights, LSBS learns its aggregation weights adaptively from the deformation features themselves. This makes the smoothing operation content-aware, enabling it to selectively retain or emphasize deformation patterns according to the local context. Consequently, LSBS not only enforces local spatial consistency but also introduces additional feature transformation capability beyond that of conventional smoothing methods.
 
\subsection{Registration Head}
To transform the deformation features produced by the Neural Demons Layer into a dense geometric transformation, we employ a lightweight Registration Head at each pyramid level. Concretely, a pair of $3 \times 3 \times 3$ convolutional layers maps the feature map to a stationary velocity field $\bm{v}^\ell \in \mathbb{R}^{\frac{H}{2^\ell} \times \frac{W}{2^\ell} \times \frac{D}{2^\ell} \times 3}$, which is subsequently trilinearly upsampled back to the original resolution.

To ensure that the topology is preserved, we represent the final deformation $\bm{\phi}$ using the Lie group exponential map, $\bm{\phi} = \exp(\bm{v})$. In line with standard practice\cite{balakrishnanVoxelmorphLearningFramework2019}, this is realized via the scaling and squaring approach. The integration process then iteratively composes the flow:
\begin{equation} 
	\bm{\phi} = \exp(\bm{v}) \approx \underbrace{\left( \dots \left( \bm{v}/2^T + \operatorname{Id} \right) \circ \dots \right)}_{T \text{ iterations}}, 
\end{equation}
where $T$ denotes the integration steps, which is set to 7. Finally, the moving image is warped using a Spatial Transformer Network as $I_w = I_m \circ \bm{\phi}$, enabling end-to-end differentiable optimization.

\subsection{Loss Function}
In our model, the core concept is that the driving force between image pairs, which indicates local misalignment, should diminish as registration progresses. To explicitly ensure this, we develop a driving force loss $\mathcal{L}_{df}$ in the feature space to measure how the driving force changes before and after warping. This loss term acts as an auxiliary objective for training the auxiliary encoder. Using ${\bm{\phi}}$ as the estimated deformation field, the moving image $I_m$ is warped to obtain $I_w = I_m \circ {\bm{\phi}}$. The feature extraction module $\mathcal{E}^\ell_\text{c}$ then operates on $I_w$ to produce the feature tensor $\mathcal{F}^{\ell}_{cw}$. In this way, after registration, we obtain two driving force tensors: $\mathcal{D}_{mf}^\ell$ and $\mathcal{D}_{wf}^\ell$.

We enforce the constraint that the driving force should decrease after registration by minimizing the following loss
\begin{equation}
	\mathcal{L}_{df}=
    \frac{1}{L} \sum_{\ell}^{L}\mathcal{L}^\ell_{df}=\frac{1}{L} \sum_{\ell}^{L} \text{ReLU}(\mathcal{D}^\ell_{wf}-\mathcal{D}^\ell_{mf}+\delta),
	\label{eq3.10}
\end{equation}
where $\delta$ is the hyperparameter of the edge threshold.

Especially if $\mathcal{D}_{wf}^\ell > \mathcal{D}_{mf}^\ell$, the ReLU function yields positive outputs for these regions. As a result, the loss function $\mathcal{L}^\ell_{df}$ increases and must be minimized during optimization. When $\mathcal{D}^\ell_{wf}<\mathcal{D}^\ell_{mf}$, this indicates that the deformation constraints required for the registration are satisfied. At this point, the ReLU outputs in these regions are clamped to 0. Unlike traditional intensity-based similarity losses, this formulation operates in the learned feature space and captures higher-order semantic and geometric mismatches. Consequently, it introduces a significant inductive bias suited to the driving-force registration method, thereby enhancing convergence stability and interpretability.

Crucially, the role of $\mathcal{L}_{df}$ is to impose explicit controllability and transparency on the auxiliary encoder. Because the encoder is trained from scratch without external supervision, it may otherwise converge to trivial or task-irrelevant representations. This loss introduces a semantic constraint that enforces a basic registration principle: the warped image must exhibit greater structural similarity to the fixed image than the original moving image. By heavily penalizing violations of this inequality, $\mathcal{L}_{df}$ ensures that the auxiliary encoder behaves as a dependable dissimilarity metric. As a result, the extracted features faithfully capture anatomical differences rather than noise, offering a stable and interpretable foundation for subsequent driving force generation.

Two other loss terms are normalized cross-correlation loss $\mathcal{L}_{NCC}$ and a smoothness regularization term $\mathcal{L}_{smooth}$. The loss function $\mathcal{L}_{NCC}$ aims to enhance the consistency of the pixel intensity between moving and fixed images as follows
\begin{equation*}
	\mathcal{L}_{NCC} = - \sum_{\bm{x}\in\Omega} \frac{\sum_{\rho}(I_{w}|_\rho - \overline{I_{w}|_\rho})(I_{f}|_\rho - \overline{I_{f}|_\rho})}{\sqrt{\sum_{\rho}(I_{w}|_\rho - \overline{I_{w}|_\rho})^2 \sum_{\rho}(I_{f}|_\rho - \overline{I_{f}}|_\rho)^2}},
	\label{ncc}
\end{equation*}
where $I_{w}|_\rho$ and $I_{f}|_\rho$ denote the local regions extracted from $I_w$ and $I_f$, respectively. $\rho$ corresponds to a 9×9×9 sliding window.

The spatial gradients of the displacement field $\bm{u}(\bm{x})$ are constrained by a diffusion regularization term $L_{\text{diff}}$, ensuring continuity and smoothness.
\begin{equation}
	\mathcal{L}_{diff}=\sum_{\bm{x}\in \Omega}\left \| \Delta \bm{u}(\bm{x}) \right \|^2,
	\label{diffusion}
\end{equation}
where $\Delta$ represents the spatial Laplace operator.

\section{Experiments}\label{sect04}

\begin{table*}[t]
	\renewcommand{\arraystretch}{1.15} 
	\centering
	\caption{Quantitative comparison of our method and other approaches on the IXI and OASIS datasets. A higher Dice similarity coefficient (DSC\%) indicates better registration accuracy. HD95 denotes the 95th percentile of the Hausdorff distance; lower values reflect more precise boundary alignment. The metric $|J_{\bm{\phi}}|\leq0$	represents the average percentage of voxels with non-positive Jacobian determinant in the deformation field, lower values indicate fewer foldings and better structural plausibility. Values in parentheses denote standard deviations; smaller deviations reflect greater stability. Efficiency is measured by inference time (Time), number of parameters (Params), and floating-point operations (FLOPs). Best results are shown in \textbf{bold}.}
	\label{tab:results}
	
	\resizebox{\textwidth}{!}{%
		\begin{tabular}{l|ccc|cccc|cccc}
			\toprule
			\multirow{2}{*}{Method} & \multicolumn{3}{c|}{Efficiency Metrics} & \multicolumn{4}{c|}{IXI Dataset} & \multicolumn{4}{c}{OASIS Dataset} \\
			\cmidrule(lr){2-4} \cmidrule(lr){5-8} \cmidrule(lr){9-12}
			& {Time(s)$\downarrow$} & {Param(M)$\downarrow$} & {FLOPs(G)$\downarrow$}
			& DSC(\%)$\uparrow$& HD95$\downarrow$ &ASD$\downarrow$ & $|J_{\bm{\phi}}|\leq0$$\downarrow$ 
			& DSC(\%)$\uparrow$ & HD95$\downarrow$ &ASD$\downarrow$& $|J_{\bm{\phi}}|\leq0$$\downarrow$ \\
			\midrule
			
			Initial & - & - & -   
			& 53.71 (5.27) & 3.568 (0.761) & 1.862 (0.362) & - 
			& 55.70 (5.57) & 3.540 (0.918) & 1.783 (0.355) & - \\
			
			SyN & 21.73 & - & -   
			& 77.20 (3.10) & 2.790 (0.460) & 0.989 (0.154) & $<$1e-6 
			& 76.67 (4.09) & 2.469 (0.774) & 0.865 (0.217) & $<$1e-6 \\
			
			ElasticDemons & 25.16 & - & -   
			& 77.52 (2.97) & 2.766 (0.477) & 0.977 (0.155) & $<$1e-6 
			& 76.85 (3.76) & 2.452 (0.748) & 0.858 (0.206) & $<$1e-6 \\
			
			\midrule
			
			VoxelMorph & \textbf{0.14} & \textbf{1.10} & 512.5 
			& 76.60 (2.62) & 2.779 (0.407) & 0.990 (0.126) & 4.48e-4 
			& 77.34 (2.90) & 1.696 (0.402) & 0.812 (0.136) & 2.90e-4 \\
			
			TransMorph & \textbf{0.19} & 46.69 & 713.5 
			& 78.55 (2.54) & 2.639 (0.398) & 0.964 (0.117) & 2.90e-4 
			& 79.87 (2.86) & 1.571 (0.341) & 0.716 (0.129) & 1.90e-4 \\
			
			XMorpher & 1.62 & 14.80 & 1355  
			& 77.00 (2.67) & 2.731 (0.390) & 0.970 (0.121) & 3.01e-4 
			& 78.53 (2.89) & 1.610 (0.385) & 0.753 (0.144) & 1.83e-4 \\		
			
			GroupMorph & 0.46 & 1.37 & 237.3  
			& 79.02 (2.41) & 2.582 (0.374) & 0.951 (0.134) & 9.71e-5 
			& 80.56 (2.38) & 1.531 (0.315) & 0.689 (0.119) & 5.18e-5 \\
			
			Vit-V-Net & 0.33 & 110.6 & 1050  
			& 76.81 (2.61) & 2.750 (0.403) & 0.977 (0.138) & 6.55e-4 
			& 78.05 (2.83) & 1.601 (0.352) & 0.733 (0.132) & 1.87e-4 \\
			
			\midrule
			
			DrivenMorph(Ours) & 0.38 & 6.23 & \textbf{151.8}  
			& \textbf{80.23 (2.27)} & \textbf{2.479 (0.322)} & \textbf{0.920 (0.103)} & \textbf{3.26e-5} 
			& \textbf{80.68 (2.18)} & \textbf{1.514 (0.286)} & \textbf{0.660 (0.107)} & \textbf{2.54e-5} \\
			\bottomrule
		\end{tabular}%
	}
\end{table*}

\subsection{Dataset}
\textbf{OASIS Brain Dataset} \cite{marcus2007open}: This dataset comprises 414 T1-weighted brain MR scans from OASIS, classified into 35 categories. Each volume received skull stripping, spatial normalization, and subcortical segmentation, maintaining a native resolution of 160 × 192 × 224. Data were randomly divided into 354 for training, 20 for validation, and 40 for testing. With mutual cross-registration, this resulted in 62,481 training pairs, 190 validation pairs, and 780 testing pairs.

\textbf{IXI Brain Dataset}: We employed 576 T1-weighted MR images from the publicly available IXI Brain dataset, preprocessed identically to OASIS (i.e. skull stripping, spatial normalization and subcortical segmentation), then uniformly cropped to dimensions of 160 × 192 × 224. Consistent with \cite{Transmorph}, the dataset was randomly split into 403 for training, 58 for validation, and 115 for testing.

\textbf{LiTS Dataset}\cite{bilic2023liver}: To assess how well our method generalizes to large-deformation abdominal CT registration tasks, we employed 180 scans from the Liver Tumor Segmentation Challenge (LiTS). Each 3D volume was preprocessed using the corresponding segmentation masks to extract the region of interest, then cropped to a size of $192 \times 192 \times 96$. We split the dataset into 155 subjects for training and 25 for testing, resulting in 11,935 training pairs and 300 testing pairs. Registration accuracy was quantified using the Dice overlap computed on the liver segmentations.

\subsection{Implementation Details}
We evaluated our method against representative baselines, including traditional optimization-based algorithms (SyN\cite{avants2008symmetric}, ElasticDemons\cite{zikic2011general}) and learning-based models (VoxelMorph\cite{balakrishnanVoxelmorphLearningFramework2019}, TransMorph\cite{Transmorph}, XMorpher\cite{shi2022xmorpher}, GroupMorph\cite{tan2024groupmorph}, Vit-V-Net\cite{chenViTVNetVisionTransformer2021}). Traditional methods were executed on a workstation with an Intel Core i7-12700 CPU, while learning-based models were implemented in PyTorch 2.3.1 and trained on an Ubuntu system with an NVIDIA GeForce RTX 3090 GPU.

All models were trained for 300 epochs with a batch size of 1, using the Adam optimizer ($lr=10^{-4}$) together with a ReduceLROnPlateau learning rate scheduler. To ensure a fair comparison, we set the loss weights to $\lambda_{NCC}=1$ and $\lambda_{diff}=1$ for all methods. In addition, our method incorporated an auxiliary force loss term with a weight of $\lambda_{df}=20$. The local search radius for the neighborhood correlation was set to $r=2$.

\begin{figure*}[t]
	\centerline{\includegraphics[width=0.8\linewidth]{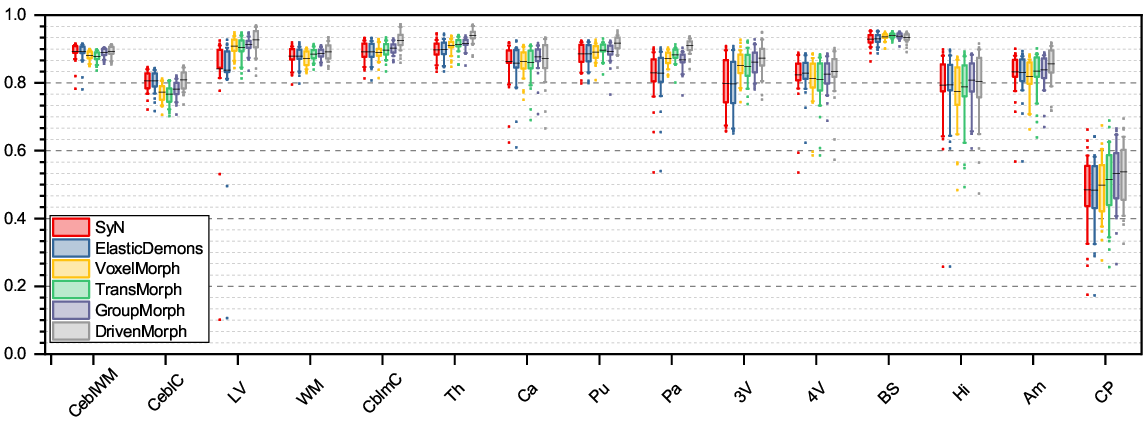}}
	\caption{
		Boxplots that illustrate the Dice coefficient of 15 anatomical structures on the OASIS dataset. The cerebral white matter (CeblWM), cerebellum cortex (CblmC), lateral ventricle (LV), cerebellum white matter (WM), cerebellum cortex (CblmC), thalamus (Th), caudate (Ca), putamen (Pu), pallidum (Pa), 3rd ventricle (3V), 4th ventricle (4V), brain stem (BS), hippocampus (Hi), amygdala (Am), and choroid-plexus (CP) are included.}
	\label{fig4_2}
\end{figure*}

\subsection{Experimental Analysis}

\begin{figure*}[htbp]
	\centerline{\includegraphics[width=0.93\linewidth]{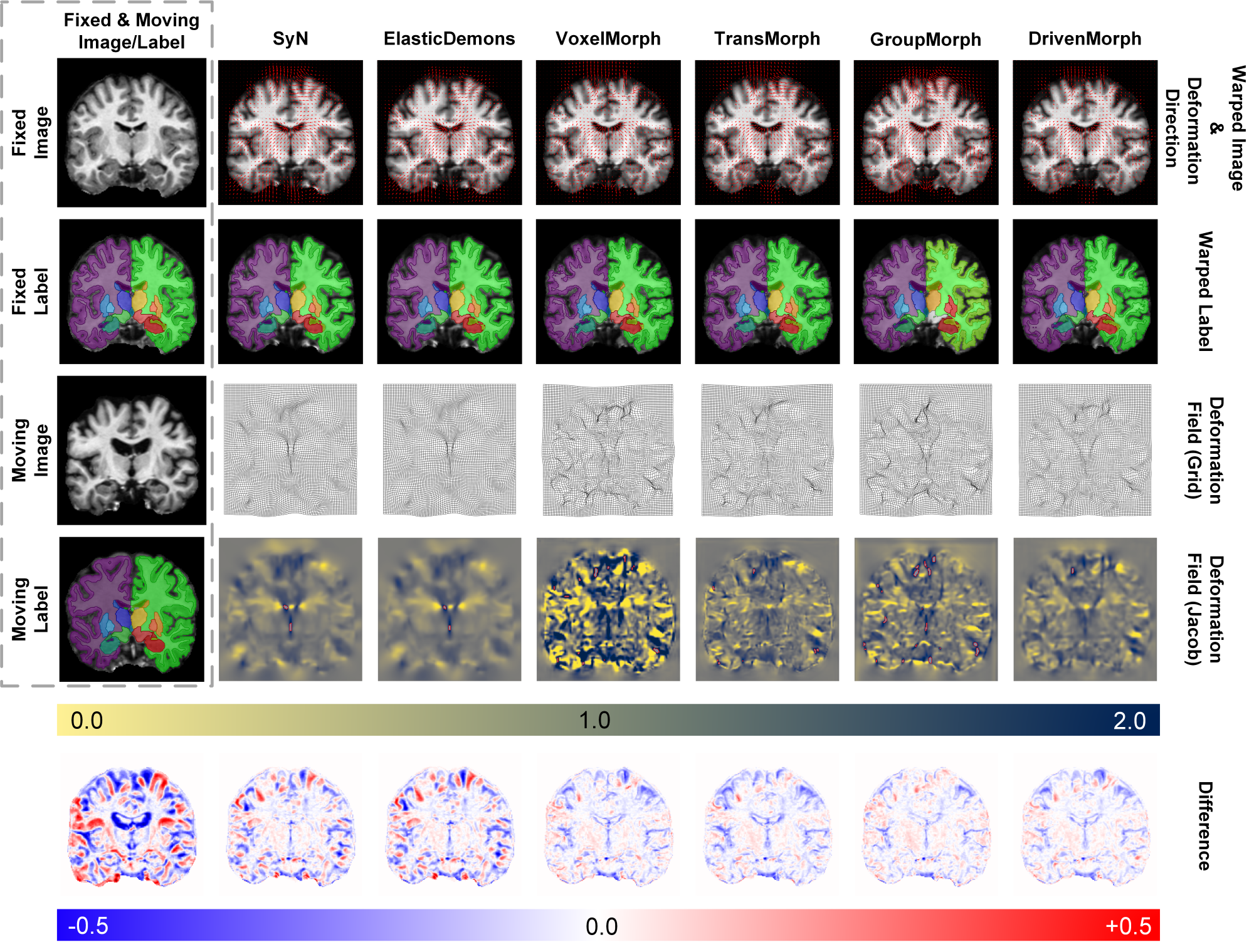}}
	\caption{Qualitative comparison of registration results on the OASIS dataset using different methods. The first column displays the fixed image, its segmentation label, the moving image, its segmentation label, and the absolute difference between the fixed and moving images. From left to right (excluding the first column): results from SyN, ElasticDemons, VoxelMorph, TransMorph, GroupMorph, and our proposed method (DrivenMorph). From top to bottom: the deformed moving image overlaid with the deformation direction, the deformed segmentation label, the deformation field visualized as a grid, the Jacobian determinant of the deformation field (The area surrounded by the red outline is the folding area), and the absolute difference between the deformed image and the fixed image.}
	\label{fig4_1}
\end{figure*}

\subsubsection{Comparison with Traditional Methods}
Table \ref{tab:results} presents the quantitative registration performance on the IXI and OASIS datasets. The proposed DrivenMorph consistently outperforms traditional optimization-based methods (SyN and ElasticDemons) in overlap accuracy (DSC), boundary precision (HD95, ASD), and overall registration quality, while maintaining competitive topological regularity, as indicated by a low folding rate.
On the IXI dataset, our DrivenMorph achieves a Dice score of 80.23\%, representing a relative improvement of 3.5\% over ElasticDemons and 3.9\% over SyN. It also reduces HD95 by 10.4\% and 11.2\%, respectively, and achieves the lowest ASD, with a 5.8\% reduction over the best-performing traditional baseline, indicating enhanced anatomical alignment and boundary delineation. On the OASIS dataset, the proposed DrivenMorph yields a 5.2\% Dice improvement over SyN, along with HD95 and ASD reductions of 38.7\% and 23.0\%, respectively, further demonstrating its robustness across datasets.
Qualitative results in Fig. \ref{fig4_1} corroborate these results. Compared to traditional methods, the proposed DrivenMorph produces more homogeneous, lower-intensity difference maps, reflecting reduced misalignment and improved correspondence with the fixed image. While SyN and ElasticDemons exhibit near-zero folding rates ($<1\mathrm{e}{-6}$), DrivenMorph also maintains low folding rates ($3.26\mathrm{e}{-5}$ on IXI, $2.54\mathrm{e}{-5}$ on OASIS), indicating that its learned deformation fields are both topologically stable and anatomically plausible.

As shown in Fig. \ref{fig4_2}, traditional methods outperform learning-based models in cerebellar regions such as Cblwm and Cblc structures with large volumes, stable intensities and clinical significance, where classical optimizations are typically well-tuned. Notably, our DrivenMorph achieves comparable accuracy in these areas while exhibiting greater consistency across less prominent structures, where traditional methods often degrade due to lack of tailored priors. This highlights the robustness and broader applicability of our data-driven approach.

\subsubsection{Comparison with Deep Learning Methods}
We also compare the proposed DrivenMorph with several popular deep learning-based registration models. On the IXI dataset, DrivenMorph achieves the highest Dice score (80.23\%) among all learning-based methods, surpassing the previous best by 1.21\%. It also consistently achieves the lowest HD95 (2.479) and ASD (0.920), indicating improved anatomical overlap and more precise boundary alignment. Similarly, on the OASIS dataset, DrivenMorph maintains its lead with a Dice score of 80.68\% and achieves the lowest values of HD95 (1.514) and ASD (0.660) in all methods. 

We observe that GroupMorph already serves as a strong baseline and that the numerical improvements in overlap measures are comparatively modest. However, these improvements are achieved together with significantly better deformation regularity and enhanced computational efficiency. As shown in Table \ref{tab:results}, on the OASIS dataset, DrivenMorph lowers the proportion of non-positive Jacobian determinants to roughly half that of GroupMorph, indicating substantially fewer foldings and better preservation of topology. In addition, by explicitly modeling and supervising the driving force underlying the deformation, DrivenMorph provides a more transparent and structurally interpretable registration mechanism, enabling direct inspection of how and why local deformations arise.

As shown in Fig. \ref{fig4_1}, qualitative comparisons further validate the advantages of DrivenMorph. In difference maps, GroupMorph and DrivenMorph exhibit more uniform intensity distributions compared to other learning-based methods, indicating better registration accuracy. From a visual point of view, the differences between these two models are subtle in this representation.
However, when examining the deformation fields (grid representations), the proposed DrivenMorph produces significantly smoother and more coherent transformations. Its deformation pattern appears more regular and structurally consistent, resembling the output of traditional optimization-based methods such as SyN and ElasticDemons. This is further corroborated by the Jacobian determinant maps, where DrivenMorph demonstrates greater uniformity and fewer irregularities, suggesting better-preserved topology and smoother deformation behavior. These observations underscore DrivenMorph's ability to combine the adaptability of deep learning with the stability characteristics traditionally seen in classic algorithms.

\subsubsection{Computational Efficiency Analysis}
Table \ref{tab:results} summarizes the computational efficiency used to evaluate practical deployability. In terms of model complexity, DrivenMorph contains only 6.23 M parameters, corresponding to a substantial reduction of about 87\% relative to TransMorph (46.69 M) and 94\% relative to ViT-V-Net (110.6 M).

Notably, even with the addition of an auxiliary encoder, our approach achieves the lowest computational cost at 151.8 GFLOPs. This theoretical cost is markedly lower than that of competing methods, amounting to just 29.6\% of VoxelMorph (512.5 G) and 21.3\% of TransMorph (713.5 G). This efficiency largely arises from our resolution management strategy: the encoder applies early spatial downsampling, so that the computationally intensive decoder operates in a lower-resolution latent space (reducing the voxel count by a factor of 8 at the initial stage). In addition, the use of sparse Local attention prevents the quadratic complexity associated with global transformer architectures.

In terms of inference speed, DrivenMorph registers a volume in roughly 0.38 seconds. Although it is slightly slower than the simplest CNN baseline, VoxelMorph (0.14 s), it outperforms other high-accuracy approaches such as GroupMorph (0.46 s). It thus comfortably satisfies the sub-second latency constraints for real-time clinical usage, offering a favorable compromise between high-precision registration and computational efficiency.

\subsection{Ablation Experiments and Interpretability Analysis}
We conducted an analysis of the interpretability of the model and the impact of the core parameters.

\subsubsection{Directional Consistency Analysis} We assess the effectiveness of the proposed neighborhood descriptors in the latent feature space by examining the directional consistency between the derived driving force $\mathcal{D}^\ell$ and the predicted deformation field. This experiment tests the hypothesis that, at each resolution scale, the learned driving direction aligns with the resulting displacement vectors, thus guiding the deformation process.

\begin{figure}[htbp]
	\centerline{\includegraphics[width=\linewidth]{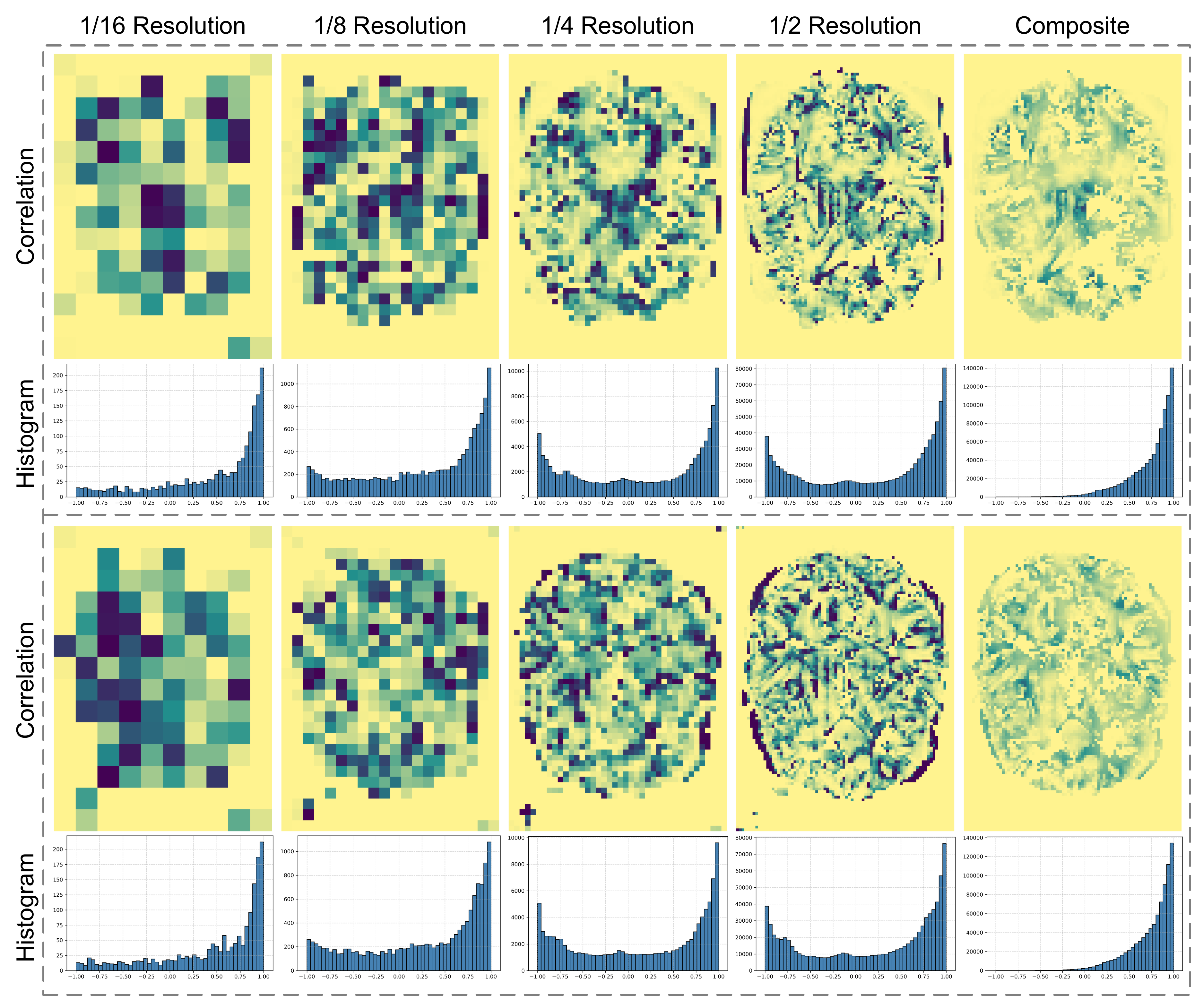}}
	\caption{Directional Consistency Between Predicted Deformation and Feature-Derived Driving Directions. Heatmap colors range from modena(low similarity) to faint yellow (high similarity), indicating the degree of directional alignment.}
	\label{fig4_3}
\end{figure}

In Fig. \ref{fig4_3}, we estimate a local driving force direction at each voxel by calculating a weighted average of discrete 3D offsets within $[-r, r]^3$, weighted by $\mathcal{D}^\ell(\bm{x})$. This results in a single 3D vector that approximates the dominant local motion trend, used exclusively for visual analysis. The cosine similarity between this estimated vector and the predicted deformation vector $\mathcal{V}^\ell(\bm{x})$ measures their directional agreement. This procedure is repeated across all pyramid levels $\ell$, producing per-scale heatmaps and global histograms to fuse the final result; therefore, a composite heatmap highlights the maximum similarity across scales.

Importantly, the increase in irrelevant (low-similarity) regions at high resolutions does not indicate reduced performance. Instead, it demonstrates the complementary construction of the framework: coarse levels manage global displacements, while finer levels serve as corrective modules, refining or reversing deformations to enhance local accuracy. This complementary nature is apparent in the composite visualization, where the integration across scales ensures optimal directional consistency.

\subsubsection{Analysis of Search Radius in Driving Force}
\begin{figure}[htbp]
	\centerline{\includegraphics[width=\linewidth]{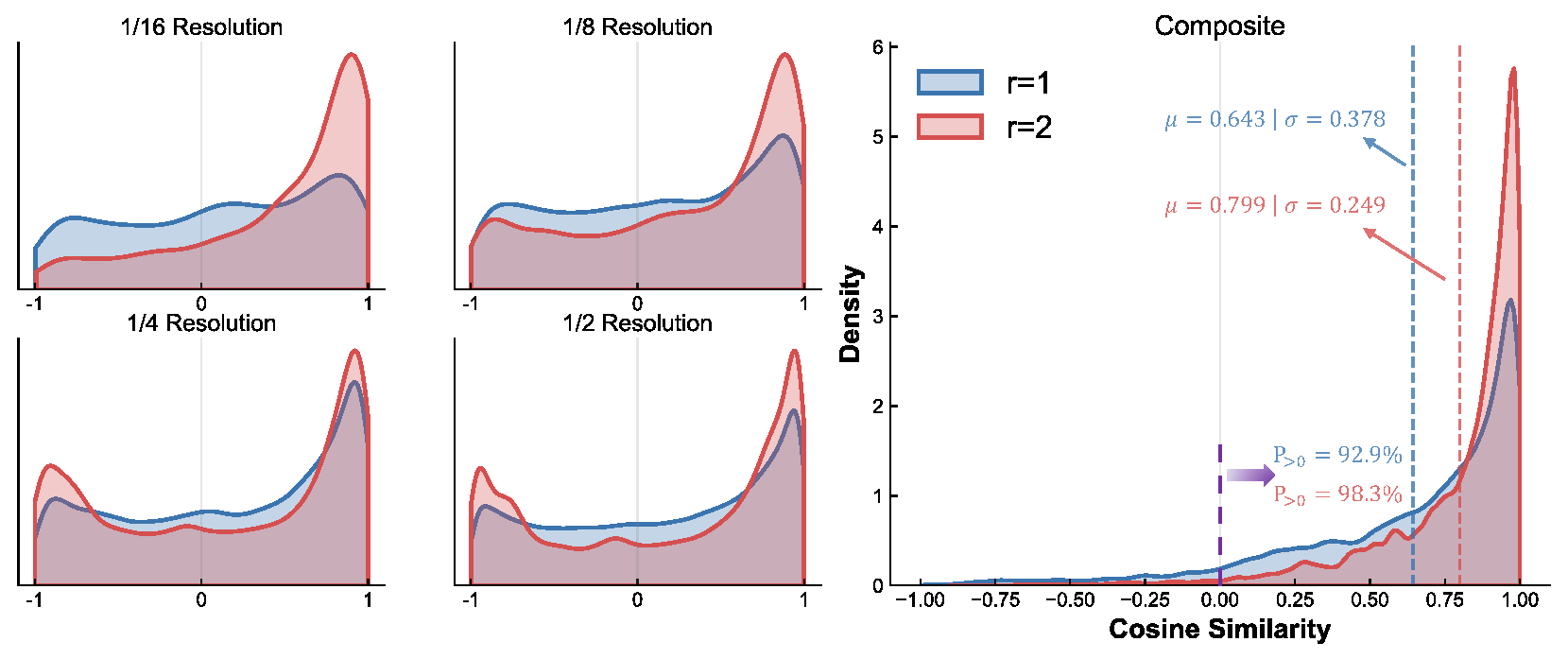}}
	\caption{Quantitative analysis of directional consistency distributions between displacement fields and driving forces under different search radii. The plots display cosine similarity densities for search radii $r=1$ (blue) and $r=2$ (red) across multi-scale residual displacement fields (Left) and the final composite field (Right). Statistical metrics ($\mu, \sigma, P_{>0}$) are annotated to quantify the alignment. A higher $\mu$ and $P_{>0}$ with a lower $\sigma$ (as observed in $r=2$) indicate a stronger and more concentrated directional consistency.}
	\label{fig4_4}
\end{figure}

Table \ref{tab:r} outlines how varying the search radius $r$ of the driving force affects registration performance. Increasing $r$ enlarges the local neighborhood for aggregating driving-related features, thus providing richer and globally consistent guidance, which improves Dice scores across both datasets. However, this also increases folding rates because of the inclusion of noisy or conflicting local signals, leading to sharper, irregular deformations. Thus, while a larger search radius enhances the driving potential of $\mathcal{D}^\ell$, it may also cause more dramatic deformations. This phenomenon is visually supported by Fig. \ref{fig4_4}, where larger radii produce more peaked consistency distributions (decreasing $\sigma$ from 0.378 to 0.249), indicating stronger directional consensus (increasing $\mu$ from 0.643 to 0.799), whereas smaller radii result in flatter distributions, reflecting weaker but more uniformly distributed driving tendencies.

Overall, the search radius $r$ acts as a structural hyperparameter that negotiates the trade-off between local specificity and directional stability. In practice, we find that intermediate settings ($r=2$) yield a reliable compromise between accuracy and smoothness (achieving a high positive directional rate of $P_{>0}=98.3\%$), without requiring fine-grained tuning.

\begin{table}[htbp]
\centering
\caption{The influence of different Radius $r$ of driving force}
\label{tab:r}
\begin{tabular}{ccccc}
\toprule
$ r $
& \multicolumn{2}{c}{OASIS} & \multicolumn{2}{c}{IXI} \\
\cmidrule(lr){2-3} \cmidrule(lr){4-5}
& DSC & $|J_{\bm{\phi}}|\leq0$ & DSC & $|J_{\bm{\phi}}|\leq0$ \\
\midrule
0 & 73.93(3.91) & 2.91e-6 & 74.16(3.24) & 3.61e-6 \\
1 & 77.67(2.98) & 9.04e-6 & 78.13(2.58) & 1.03e-6 \\
2 & 80.68(2.18) & 2.54e-5 & 80.23(2.27) & 3.26e-5 \\
\bottomrule
\end{tabular}
\end{table}

\subsubsection{Ablation Study on Driving Force Loss}
We conducted an ablation study to examine the effects of adjusting the weight of the driving force loss $\mathcal{L}_{df}$. As illustrated in Table \ref{tab:contrastive}, incorporating $\mathcal{L}_{df}$ is the main factor behind the performance gains, with a substantial improvement observed when increasing $\lambda$ from $0$ to $10$. The model's performance remains stable for $\lambda \in [10, 30]$. This indicates that the driving force loss functions primarily as a structural constraint rather than a finely tuned hyperparameter. We therefore fixed $\lambda=20$ in all experiments, which provided a reliable compromise between accuracy and regularization across the various datasets.

\begin{table}[htbp]
\centering
\caption{Ablation Study on Driving Force Loss Weight}
\label{tab:contrastive}
\begin{tabular}{ccccc}
\toprule
$\mathcal{L}_{df}$
& \multicolumn{2}{c}{OASIS} & \multicolumn{2}{c}{IXI} \\
\cmidrule(lr){2-3} \cmidrule(lr){4-5}
& DSC & $|J_{\bm{\phi}}|\leq0$ & DSC & $|J_{\bm{\phi}}|\leq0$ \\
\midrule
0 & 77.93(2.75) & 2.91e-4 & 77.16(2.65) & 3.61e-4 \\
10 & 79.67(2.45) & 9.04e-5 & 78.91(2.43) & 1.03e-4 \\
20 & 80.68(2.18) & 2.54e-5 & 80.23(2.27) & 3.26e-5 \\
30 & 79.71(2.23) & 2.91e-5 & 79.35(2.31) & 3.61e-5 \\
\bottomrule
\end{tabular}
\end{table}

\subsubsection{Ablation Study on DrivenMorph Components}
To assess the importance of the proposed latent-space driving force field and its associated modules, we perform an extensive ablation study, summarized in Table \ref{tab:ablation}. In each setting, a particular component is either removed or substituted, while the rest of the architecture remains unchanged.
	
First, we replace the auxiliary encoder with raw intensity differences ($\Delta I^\ell(\bm{x}, \bm{y}) = I^\ell_f(\bm{y}) - I^\ell_m(\bm{x})$) to construct the driving force field \eqref{driving force}, directly mirroring classical intensity-based registration. While this variant leads to a clear performance decline (DSC: 80.68 $\rightarrow$ 74.24), it still converges reliably and preserves low folding rates. This observation indicates that the Neural Demons Layer itself does not require feature fusion to operate, whereas the latent feature representation greatly enhances the expressiveness of the driving force in capturing complex anatomical differences.

Next, we replace the proposed neighborhood driving force field($Corr_{nh}$) with a Patch-based correlation ($Corr_p$). Despite its local formulation and comparable computational cost, this substitution results in a severe performance collapse (DSC: 58.31). This failure is not merely due to reduced representation capacity, but rather reflects a structural incompatibility: $Corr_p$ captures localized similarity magnitudes within a patch, it does not encode spatially aligned directional correspondence between neighboring voxels. In our framework, the Neural Demons Layer requires such spatially aligned discrepancy information to guide force-based deformation updates. This structural mismatch prevents $Corr_p$ from serving as a valid driving force representation in this context.

We further assess the contribution of the LSBS by completely removing it from the model. The observed drop in performance shows that, while the LCBD module provides useful guidance, an extra feature transformation stage is still required to translate it into rich deformation fields, which is a common design principle in deep learning.

Finally, replacing LCBD with a vanilla attention mechanism yields poorer accuracy and increased irregularity, even though the overall model complexity remains similar. This finding shows that our attention formulation is not a trivial variant, but rather a carefully constrained design that enforces locality and directional consistency as an inductive bias inspired by the Demons update rule.

The above ablation results demonstrate that the introduced components are not incremental additions, but structurally necessary elements for realizing a force-driven and interpretable deformation mechanism within a deep learning framework.

\begin{table}[htbp]
	\centering
	\caption{Ablation study of DrivenMorph components on the OASIS dataset. Each configuration investigates the necessity of a specific module by removal (w/o) or replacement with (Repl. w/).}
	\label{tab:ablation}
	\setlength{\tabcolsep}{1pt} 
	
	\begin{tabular}{lccccc}
		\toprule
		\textbf{Config.} & 
		\makecell[b]{\textbf{DSC} \\ \textbf{(\%)} $\uparrow$} & 
		\makecell[b]{\textbf{Folding} \\ $\downarrow$} & 
		\makecell[b]{\textbf{Time} \\ \textbf{(s)}} & 
		\makecell[b]{\textbf{Params} \\ \textbf{(M)}} & 
		\makecell[b]{\textbf{FLOPs} \\ \textbf{(G)}} \\ 
		\midrule
		
		Base & 80.68(2.18) & 2.5e-5 & 0.38 & 6.23 & 151.8 \\ \midrule
		Repl. w/ Intensity & 74.24(3.84) & 7.1e-6 & 0.33 & 5.40 & 93.9 \\
		Repl. w/ $Corr_p$. & 58.31(5.03) & 0 & 0.39 & 6.23 & 151.8 \\
		w/o LSBS & 77.93(3.01) & 2.9e-4 & 0.31 & 5.56 & 135.0 \\
		Repl. w/ Vanilla Attn & 79.02(2.88) & 1.9e-4 & 0.45 & 6.23 & 152.9 \\ 
		\bottomrule
	\end{tabular}

\end{table}

\subsection{Generalizability Validation}
\begin{table}[htbp]
	\centering
	\caption{Quantitative evaluation on the LiTS abdominal CT dataset. This experiment assesses the generalizability of the proposed method on large-deformation inter-subject registration tasks (Best results are shown in \textbf{bold}).}
	\label{tab:lits_results}
	\resizebox{0.48\textwidth}{!}{%
		\begin{tabular}{l|cccc}
			\toprule
			Method & DSC(\%)$\uparrow$ & HD95$\downarrow$ & ASD$\downarrow$ & $|J_{\bm{\phi}}|\leq0$$\downarrow$ \\
			\midrule
			ElasticDemons & 85.63 (5.51) & 13.37 (6.45) & 4.46 (2.09) & \textbf{0} \\
			VoxelMorph & 87.36 (4.66) & 12.40 (5.57) & 3.89 (1.68) & 2.11e-04 \\
			TransMorph & 88.79 (4.23) & 12.12 (5.86) & 3.59 (1.64) & 1.37e-03 \\
			GroupMorph & 91.40 (2.62) & 10.37 (5.30)& 2.75 (1.21) & 2.35e-04 \\
			\midrule
			\textbf{DrivenMorph (Ours)} & \textbf{92.56 (2.47)} & \textbf{9.57 (4.99)} & \textbf{2.44 (0.98)} & 3.73e-05 \\
			\bottomrule
		\end{tabular}%
	}
\end{table}

\begin{figure}[htbp]
	\centerline{\includegraphics[width=\linewidth]{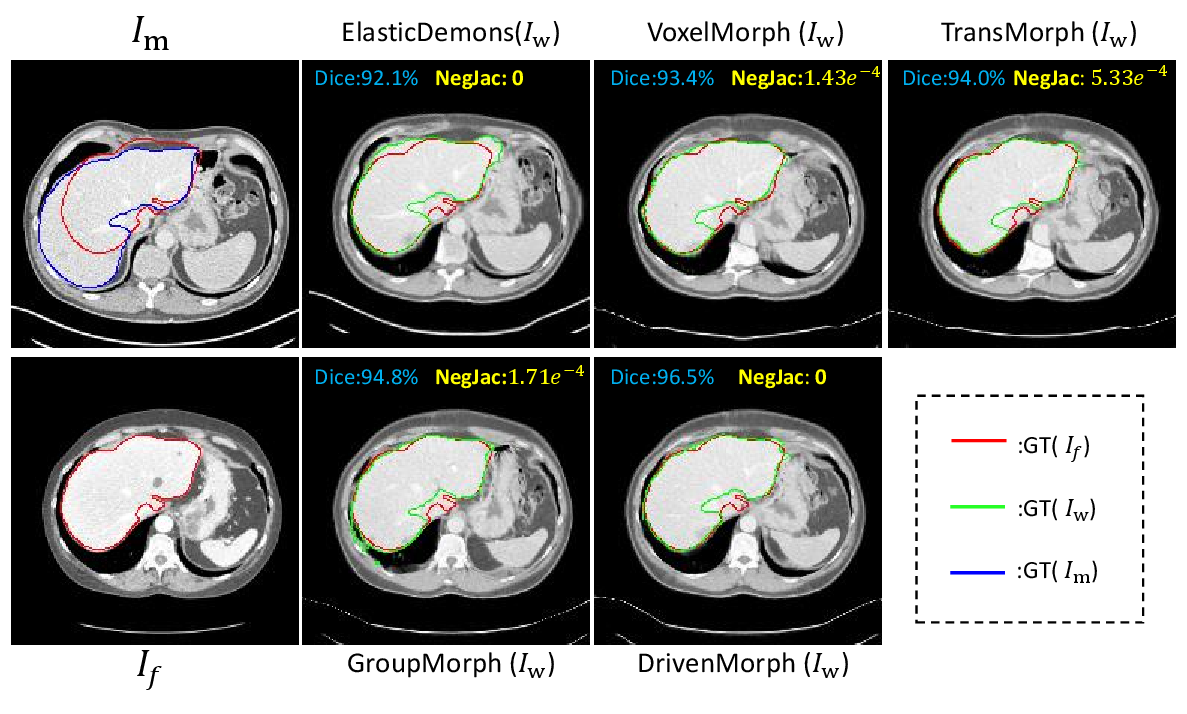}}
	\caption{Visualization of generalization performance on liver registration. The moving image ($I_m$) is registered to the fixed image ($I_f$) using various methods.  Color-coded contours show the alignment: Red ($GT(I_f)$): Target boundary; Blue ($GT(I_m)$): Initial boundary; Green ($GT(I_w)$): Prediction.}
	\label{fig4_6}
\end{figure}

To evaluate how well DrivenMorph generalizes beyond brain MRI, we further tested it on the LiTS abdominal CT dataset. Abdominal registration is particularly challenging because of the large-scale, non-linear deformations of soft tissues and the substantial anatomical variability across subjects. We compared our approach against four representative methods: the classical optimization-based ElasticDemons, the CNN-based VoxelMorph, and the transformer-based TransMorph and GroupMorph.
	
As shown in Table \ref{tab:lits_results}, DrivenMorph demonstrates superior performance in this challenging large-deformation scenario, achieving the highest registration accuracy (DSC: 92.56\%). This performance margin exposes the limitations of single-stage frameworks such as VoxelMorph and TransMorph, whose one-shot regression lacks the iterative refinement needed to handle large anatomical differences. GroupMorph narrows this gap through grouping strategies, but still underperforms compared to DrivenMorph, which reliably captures large deformations via its explicit coarse-to-fine driving force mechanism. This quantitative improvement is further illustrated in Fig. \ref{fig4_6}, where DrivenMorph achieves the closest overlap with the ground truth boundaries (red/green alignment). In addition, our method better preserves topology: large deformations often lead to folding artifacts, but our force-driven architecture maintains both high registration accuracy and the anatomical plausibility required for clinical and research applications.

\section{Conclusion}\label{sec05}
In this work, we propose a registration framework based on the Demons algorithm, where a learnable driving force tensor is derived from local feature similarities in the latent space. This tensor captures local structural differences in terms of both direction and magnitude through a neural Demons layer, modeling deformation as "magnitude × direction". Our approach enhances interpretability, modularity, and registration accuracy compared to several benchmarks. By decoupling difference modeling from deformation generation, the framework restores physical interpretability to learning-based registration, thereby improving supervision clarity and control, and paving the way for more interpretable and structured registration systems. Experiments demonstrate strong directional consistency between the learned force and the resulting deformation, facilitating intuitive visualization of the deformation process.

This perspective also highlights an important direction for further improvement. In DrivenMorph, the driving-force tensor serves as the key interface that converts feature representations into geometric updates; as a result, its reliability is therefore limited by the anatomical semantics encoded in those features. The auxiliary driving-force loss provides a form of regularization for this interface, but by itself it cannot introduce semantic correspondences that are missing from the feature space. This limitation becomes especially problematic when the desired mapping cannot be adequately modeled as a smooth deformation, for example in cases of severe pathology, surgically removed or missing tissue, or anatomies that lie far outside the training distribution. In these scenarios, the model may still attempt to explain structural absence or true non-correspondence as if they were merely deformable variations. A promising direction for future work is therefore to strengthen feature extraction using medical foundation models, enabling the driving force to exploit richer anatomical and pathological semantics and thereby improving robustness when local intensity or texture similarities are not sufficient.

\section*{References}
\bibliographystyle{IEEEtran}
%\normalem
\bibliography{refs-jbhi2026}

\end{document}